\documentclass{esannV2}
\usepackage[dvips]{graphicx}
\usepackage[latin1]{inputenc}
\usepackage{amssymb,amsmath,array,amsthm}
\usepackage{xcolor}

\newcommand{\Tensor}[1]{\ensuremath{\mathbf{ #1}}}
\newcommand{\Matrix}[1]{\ensuremath{#1}}
\newcommand{\Vector}[1]{\ensuremath{\mathbf{#1}}}

\usepackage[a4paper]{geometry}
\usepackage{fancyhdr}
\begin{document}
\title{Tensor Decompositions in Deep Learning}

\author{Davide Bacciu$^1$, Danilo P. Mandic$^2$
%
\thanks{This work has been supported by MIUR under project SIR 2014 LIST-IT (RBSI14STDE).}
%
\vspace{.3cm}\\
%
$^1$ Department of Computer Science - University of Pisa\\
Largo Bruno Pontecorvo, 3, 56127, Pisa - Italy\\
$^2$ Department of Electrical and Electronic Engineering - Imperial College\\
Exhibition Road, London, United Kingdom
}

\maketitle

\begin{abstract}
The paper surveys the topic of tensor decompositions in modern machine learning applications. It focuses on three active research topics of significant relevance for the community. After a brief review of consolidated works on multi-way data analysis, we consider the use of tensor decompositions in compressing the parameter space of deep learning models. Lastly, we discuss how tensor methods can be leveraged to yield richer adaptive representations of complex data, including structured information.  The paper concludes with a discussion on interesting open research challenges.
\end{abstract}

\section{Introduction}
In the latter years, tensor methods have been gaining increasing interest in the machine learning community.  Multiway data analysis is one of their earliest and most popular application, addressing the processing of large scale and highly complex data, such as multivariate sensor signals. A tensor can be seen as a generalization of multidimensional arrays where the traditional algebraic operators are extended accordingly (e.g. element-wise sum, inner and outer products, etc). A tensor representation allows to capture complex interactions among input features which would not be evident on {\it flattened} data \cite{Cichocki2015TensorAnalysis}. They are a flexible data structure  allowing to seamlessly reshape vectorial data to a tensor representation for multi-linear analysis (tensorization) and viceversa (vectorization).  Clearly, any analysis performed on a full tensorial representation easily results in so called curse-of-dimensionality problems, with the complexity growing exponentially with the tensor order. This is where tensor decompositions play a fundamental role, allowing to reduce the complexity of the representation space and preserving computational feasibility of the analysis,  without a drastic reduction in the ability to capture high-order relationships in the data \cite{Cichocki2015TensorAnalysis}. Tensor decompositions operate similarly to their matrix counterpart, by decomposing high-dimensional tensors into a sum of products of lower dimensional factors.

Apart from their direct application to multi-way input data analysis, tensors are widely adopted as a fundamental building block for machine learning models. Firstly, they have found application in a variety of machine learning paradigms, ranging from neural networks \cite{cpdecompCNN,NoikovTensorizing} to probabilistic models  \cite{Castellana2019}, to enable the efficient compression of the model parameters leveraging tensor decomposition methods. Secondly, they provide a means to extend existing vectorial machine learning models to capture richer data representations, where tensor decompositions provide the necessary theoretical and methodological backbone to study, characterize and control the expressiveness of the model \cite{Cichocki2017TensorPerspectives}.

This tutorial paper takes pace from the homonym special session of the {\it 28th European Symposium on Artificial Neural Networks, Computational Intelligence and Machine Learning} to provide a focused survey of the use of tensor decompositions in deep learning models and applications. Given the constraints of a short communication, we will not provide a detailed introduction to tensor methods, whereas we will focus on reviewing three interesting and broad research topics, following the themes of the contributions presented in the session. An interested reader will find in literature comprehensive survey works introducing tensor decompositions \cite{Kolda2009}, their applications to data analysis \cite{Cichocki2017TensorPerspectives} and machine learning \cite{surv2019}. The remainder of the paper is organized as follows: Section \ref{sect:back} provides a brief background on tensors and notable decompositions.  Section \ref{sect:multiway} reviews {\it classical} applications of tensor decompositions to multi-way signal processing and data analysis. Section \ref{sect:compress} discusses more works leveraging decompositions in neural model compression, while Section \ref{sect:aggregate} focuses on the relevant topic of tensor methods as enabler for learning more expressive representations of complex data, including structured samples such as trees, networks and graphs. The paper concludes with a discussion of interesting open challenges in the field.

\section{Tensor decomposition overview} \label{sect:back}
In the following, we provide a brief introduction to tensors and their terminology. We also present three popular tensor decompositions that are representative of three classes of methods characterized by different assumptions and aimed to capture different types of relationships in multi-way data. For a more in-depth introductions readers are referred to \cite{Kolda2009} and \cite{Cichocki2015TensorAnalysis}.

\subsection{Tensors: definition and terminology}
A tensor describes a multilinear relationship between algebraic objects which span from a scalars to tensors themselves. Its definition can be given based on different mathematical concepts. For the sake of simplicity, we introduce tensors based on their intuitive view as a a multi-dimensional array. In this context, a tensor $\Tensor{A} \in \mathbb{R}^{n_1 \times \dots \times n_d}$ is an element of the tensor product of $d$ vector spaces, such that the corresponding multi-dimensional array is $\Tensor{A}(i_1,\dots,i_d) \in \mathbb{R}, \; i_k \in [1,n_k]$ where:
\begin{itemize}
	\item $i_k$ represents the index along the $k$-th dimension, known as {\it mode};
	\item $d$ is the {\it order} of $\Tensor{A}$, i.e. the number of modes;
	\item $n_k$ represents the {\it size} along the $k$-th mode.
\end{itemize}
Subarrays can be extracted from the full tensor by fixing a subset of the $i_k$ indices. Notable subsarrays are the {\it fibers}, which are the multi-way equivalent of vectors and are defined by fixing every index but one, and the {\it slices}, that are defined by fixing all but two
indices. The {\it norm} of a tensor is the square root of the sum of the
squares of all its elements, i.e. analogous to a matrix Frobenius norm. The {\it inner product} of two same-sized tensors is the sum of the element-wise products of their entries. The {\it tensor product} is the generalization of the vector outer product concept to tensors, i.e. through the outer product of $d$ vectors we can obtain the tensor $\Tensor{A} \in \mathbb{R}^{n_1 \times \dots \times n_d}$:
\[
  \Tensor{A} = \Vector{a}_1 \odot \Vector{a}_2 \odot \dots \odot \Vector{a}_d
\]
where $\Vector{a}_k$ is the $n_k$-dimensional vector corresponding to the $k$-th mode in the tensor. A relevant concept is that of {\it rank-1} tensor, that is a $d$-way object which can be strictly decomposed as the tensor product of $d$ vectors.
A tensor decomposition expresses a tensor in terms of a sequence of sum and products operating on simpler multi-way objects. In the following we summarize three relevant and popular methods heavily inspired by matrix decomposition.

\subsection{Canonical Polyadic (CP) decomposition}
The CP decomposition \cite{cpd} is a representative of the family of rank decompositions, that seek to express a tensor as the sum of a finite number of rank-1 tensors. There are several analogous formulations for the CP (see \cite{Kolda2009} for a detailed account), the most general being
\begin{equation}
\Tensor{A} = \sum_{r=1}^R \lambda_r \Vector{a}_1(r) \odot \Vector{a}_2(r) \odot \dots \odot \Vector{a}_d(r)
\label{eq:CP-decomp}
\end{equation}
where $\Vector{a}_1(r) \odot \Vector{a}_2(r) \odot \dots \odot \Vector{a}_d(r)$ is the $r$-th rank-1 tensor and $\lambda_r \in \mathbb{R}^{R}$.  The value of $r$ is called \textit{canonical rank}.

\subsection{Tucker decomposition (TD)}
The TD \cite{Tucker1966} is a representative of the family of Higher-Order Singular Value Decompositions (HOSVD), that seek to find the components that best capture the variation in single modes independently one-another. The TD decomposes a tensor into a {\it core tensor}, defining how the different tensor items interact and are mixed with each other, and multiple {\it modes matrices}. Given a tensor  $\Tensor{A} \in \mathbb{R}^{n_1 \times \dots \times n_d}$, the TD of its $(i_1,\dots,i_d)$ entry is
\begin{equation}
\Tensor{A}(i_1,\dots,i_d) = \sum_{r_1=1}^{R_1} \dots \sum_{r_d=1}^{R_d} \Tensor{G}(r_1,\dots,r_d) \Matrix{U_1}(i_1,r_1)\dots \Matrix{U_d}(i_d,r_d),
\label{eq:HOSVD-decomp}
\end{equation}
where $\Tensor{G} \in \mathbb{R}^{r_1 \times \dots \times r_d}$ is the core tensor and $\Matrix{U_k} \in \mathbb{R}^{n_k \times r_k}$ are the mode matrices. The values $R_k$ denote the rank along the $k$-th dimension.

\subsection{Tensor Train (TT) decomposition}
The TT \cite{Oseledets2011} is a representative of those decompositions assuming that mode ordering is somehow semantically relevant and should be taken into consideration in the factorization. The TT approximates a tensor $\Tensor{A} \in \mathbb{R}^{n_1 \times \dots \times n_d}$ through a sequence of order-3 tensors, each connected to its left and right neighbor in the mode ordering. More formally, the TT of the $(i_1,\dots,i_d)$ entry of $\Tensor{A}$ is
\begin{equation}
\Tensor{A}(i_1,\dots,i_d) = \sum_{r_0=1}^{R_0} \dots \sum_{r_d=1}^{R_d} \Tensor{G}_1(r_0,i_1,r_1) \Tensor{G}_2(r_1,i_2,r_2) \dots \Tensor{G}_d(r_{d-1},i_d,r_d),
\label{eq:TT-decomp}
\end{equation}
where $\mathbf{G}_k \in \mathbb{R}^{R_{k-1} \times n_k \times R_k}$ are tensors. The value $R_k$ represents the TT-rank along the $k$-th mode and it is such that $R_0 = R_d = 1$.

\section{Multi-way data analysis} \label{sect:multiway}
Among the different uses of tensor decompositions in machine learning, multi-way data analysis has  certainly been the first to develop and, yet, the most popular \cite{Cichocki2015TensorAnalysis}.
By multi-way analysis, we refer to the fact that tensor decompositions are used as the adaptive method to process, organize and make sense of the available data and its complex relationships, without resorting to {\it external} learning models. In this brief survey, we focus on those data analysis applications which, for complexity and scale, are typically within the scope of deep learning applications.

A type of data relationship that is often addressed through tensor decompositions is the sequential one, i.e. where the data collection comprises samples which follow a complete order. A typical example is timeseries data: this is typically a 3-way tensor in which slices represent a snapshot of the data at a given time. Here, tensor decomposition allows to isolate latent structures in the data: \cite{tempCP}, for instance, leverages CP to discover patterns in the temporal evolution of publishing activities and to perform look-ahead predictions, while in \cite{anomalCP} CP is used for detecting anomalies.

One of the works presented in the special session falls into this line of research. In \cite{session1}, the authors discuss an application of tensor decomposition on short text game commentary data to understand the temporal changes in the effectiveness of cricket players. To this end, the paper shows the use of a Tucker decomposition for discrete data. The paper also releases the original data collected and used for the analysis.

Tensor decompositions can be applied to more general forms of relational data. For instance, samples characterized by the presence of different types of relationship (e.g. friendship, content sharing, tagging in social networks) can be straightforwardly tensorized by modeling each relationship in terms of the two subjects involved and the type or relation binding them, having a different tensor slice to represent each of the available relations. In \cite{Weber2017EventRW}, for instance, it is presented a natural language understanding application, where tensors are used to capture multiplicative interactions combining predicate, object and subject generating aggregated representations for event prediction tasks. In \cite{relat1}, a Tucker-like decomposition is exploited on three-way tensors to perform collective learning on network data. The use of tensor methods in graph data processing appears a lively research field, in particular as regards the prediction of new relations in knowledge graphs, where tensor decomposition show their advantages in the trade-off between expressivity and computational efficiency \cite{knowlgraph1,knowlgraph2}.

Another successful application of multi-way analysis is in information fusion: for instance,   \cite{ben2017mutan} uses tensors to fuse visual and textual representations. Overall, the applications discussed in this section seem to share a general limitation, that is related with the use of tensors with a fixed number of modes, that rarely exceeds three.

\section{Neural model compression} \label{sect:compress}
Tensor decompositions have a second major application to machine learning  which is related to their compression ability, in particular when considering the typical large-scale of data in deep learning. The relationship of tensors with the deep learning world are particularly evident when considering {\it tensor networks} \cite{Cichocki2017TensorPerspectives} that provide a general framework for representing the factorization of large tensors into networks of smaller elements. Tensor networks provide the means to effectively regulate the trade-off between the number of model parameters and the predictive accuracy. Further, they put forward a methodological framework that allows assessing the expressive power of the compressed neural models. The typical approach to compress whole deep learning architectures is discussed in detail by \cite{Cichocki2017TensorPerspectives}. In brief, given an uncompressed deep neural network (DNN), one can construct the corresponding tensor network representation. This can be then simplified, by the decompositions discussed in Section \ref{sect:back}, to achieve the desired trade-off between parameterization and predictive accuracy. Finally, the compressed tensor network is mapped back into the corresponding compressed DNN. This approach has been successfully applied to Restricted Boltzmann Machines \cite{rbmTN} and convolutional architectures \cite{pmlr-v48-cohenb16,indbiasTN}.

A different approach to neural model compression leverages tensor decomposition on the single layers of the network. For instance, \cite{NoikovTensorizing} proposes to efficiently store the dense weight matrices of the fully-connected layers of a VGG network by leveraging Tensor
Train factorization. Conversely, \cite{DBLP:Calvi} introduces the Tucker Tensor Layer as an alternative to the dense weight-matrices of neural networks. They show how to leverage tensor decomposition to drastically reduce the number of parameters in the neural layer, also deriving a specialized form of back-propagation on tensors that preserves the physical interpretability of Tucker decomposition and provides an insight into the learning process of the layer.

An alternative perspective over the use of tensor factorization in neural layers is provided by \cite{yang2017deep}, which introduces a multitask representation learning framework leveraging tensor factorisation to share knowledge across tasks in fully connected and convolutional DNN layers.

\section{Higher-Order Representation Learning} \label{sect:aggregate}
Our brief overview concludes with the most recent and, possibly, yet less developed research topic binding deep learning models and tensor decompositions. This topic mixes the application of multi-way analysis to relational data discussed in Section \ref{sect:multiway} with the use of tensors as elements of the neural layer reviewed in Section \ref{sect:compress}. In particular, we consider how tensors can be used to modify input aggregation functions inside the artificial neuron. Input aggregation in neurons is typically achieved by a weighted sum of the inputs to the unit which, for vectorial data, is equivalent to the inner product between a the weight vector and the input vector.  When dealing with tree structured data, this process in generalized in such a way that, given a specific node in the tree, the neuron recursively computes its activation by a weighted sum of the activations of its children, with appropriate weight sharing assumptions. Such an aggregation function can be easily tensorized, as postulated already by \cite{Frasconi1998}, to capture higher-order relationships between the children encodings.

The higher-order recursive neuron by \cite{Frasconi1998} has long been only a theoretical model, formulated solely for binary trees. In \cite{Bacciu2012,Bacciu2013,Bacciu2013CompositionalModel}, the model has been extended to a probabilistic formulation for general $n$-ary trees which clearly highlights the tensorial nature of the input aggregation function (that is a $n$-way tensor map). Nonetheless, for computational tractability issues, the same works approximated the tensor with a simple probabilistic factorization largely equivalent to a weighted sum in neural models. Only recently, \cite{Castellana2019} has introduced a proper tensor decomposition of the $n$-way probabilistic tensor leveraging a Bayesian Tucker decomposition.

On the side of neural models, \cite{Socher2013} discusses the use of a high-order neural network for structured data that leverages a full 3-way tensor for aggregating children information in binary parse trees within a natural language processing application. The second paper \cite{session2} in this special session, instead, introduces what is seemingly the first tensor recursive neurons for $n$-ary trees. More importantly, it proposes the use of tensor decompositions as a viable and expressive trade-off between the simplicity of sum aggregation and the complexity of full tensor aggregation. Two input accumulation functions are discussed in the paper: one leveraging CP decomposition and the other relying on the Tensor-Train factorization. Similarly, \cite{tuckerRecursive2020} shows how a tensor recursive neuron can exploit the Tucker decomposition to control the trade-off between the size of the neural representation and the size of the neural parameter space.

\section{Conclusions and Research Challenges} \label{sect:conclusion}

The use of tensor decompositions as a fundamental building block in deep learning models is a growing research topic which is, yet, in its infancy. Throughout our brief review, we have highlighted that the use of tensor factorization as a stand-alone method for multi-way data analysis is seemingly the most mature research topic. At the same time, tensor decompositions are starting to be heavily used for compression purposes in neural models, throughout well-grounded approaches that allow to compress full networks while maintaining performance guarantees, or by tensorization of full neural layers. Lastly, we have singled out a very recent and promising research direction leveraging tensor decompositions as input aggregation functions for building higher-order neurons for structured data processing, to learn more expressive neural representations of structured information.

The research themes discussed above stimulate interesting research challenges which can help
increasing the effectiveness of deep learning models and deepen our understanding of their inner workings. Tensorized neural layers put forward a research question about whether tensorization should only affect the forward phase of neural models (i.e. computing of the neural activation) or if it has non-trivial reflections also on the backward phase (i.e. learning). In this respect, \cite{DBLP:Calvi} began to discuss how Tucker decomposition can be leveraged in the backward phase to enhance interpretability of propagated gradients. Along these lines, it would be interesting to study if tensors can be leveraged to define novel weight optimization algorithms and whether they can contribute to the discussion concerning dynamics and convergence of stochastic gradient descent \cite{soattoSGD}. A second research challenge relates to representation learning with topology varying graphs. Literature reports wide use of tensors for multirelational data analysis on single networks and, as discussed in our review, some early works dealing with neural processing of collections of tree structured samples. What seems yet missing is their use in Deep Graph Networks (DGN) \cite{bacciu2019gentle,cgmm} that can handle the most general case of datasets of graph samples with unconstrained topology. It seems likely that tensorization of the graph neurons would increase their ability in capturing higher-order structural dependencies between the nodes in the graphs. This might have a very practical impact on the predictive performance of the DGN which, as shown in \cite{iclr19}, appears often lower than that of dummy baseline models. On the other hand, it would also be interesting to study how tensorization can affect the expressivity of DGN from a theoretical perspective \cite{how-powerful-gnn,esann20Errica}.

A key aspect to promote research on tensor decomposition in deep learning is the availability of software libraries integrating tensor methods within the deep learning development frameworks. While there are several consolidated libraries providing stable implementations of tensor decompositions, these are typically distributed as packages of popular scientific computing environments, such as R, Matlab and Mathematica, which are less used by the deep learning community. Nonetheless some contributions are beginning to appear also on this respect: {\tt scikit-tensor} \cite{scikit} integrates some consolidated tensor decomposition (Tucker, CP and the like) into the {\tt scikit-learn} universe. More recently, {\tt TensorD} \cite{tensord} provides a Python tensor library built on {\tt Tensorflow} and providing basic tensor operations and decompositions with support for parallel computation (e.g. GPU). {\tt TensorLy} \cite{tensorly} is a Python library implementing a wide range of methods for tensor learning, allowing to leverage different computation back-ends including {\tt NumPy}, {\tt MXNet}, {\tt PyTorch}, {\tt TensorFlow}, and {\tt CuPy}. {\tt HOTTBOX}  \cite{hottbox} is a recent standalone Python
toolbox for tensor decompositions, statistical analysis, visualisation, feature extraction, regression and non-linear classification of multi-dimensional data.


\begin{footnotesize}

\end{footnotesize}



\begin{thebibliography}{10}

\bibitem{Cichocki2015TensorAnalysis}
Andrzej Cichocki, Danilo Mandic, Lieven De~Lathauwer, Guoxu Zhou, Qibin Zhao,
  Cesar Caiafa, and Huy~Anh Phan.
\newblock {Tensor decompositions for signal processing applications: From
  two-way to multiway component analysis}.
\newblock {\em IEEE Signal Processing Magazine}, 32(2):145--163, 2015.

\bibitem{cpdecompCNN}
Vadim Lebedev, Yaroslav Ganin, Maksim Rakhuba, Ivan~V. Oseledets, and Victor~S.
  Lempitsky.
\newblock Speeding-up convolutional neural networks using fine-tuned
  cp-decomposition.
\newblock In Yoshua Bengio and Yann LeCun, editors, {\em 3rd International
  Conference on Learning Representations, {ICLR} 2015, San Diego, CA, USA, May
  7-9, 2015, Conference Track Proceedings}, 2015.

\bibitem{NoikovTensorizing}
Alexander Novikov, Dmitrii Podoprikhin, Anton Osokin, and Dmitry~P Vetrov.
\newblock Tensorizing neural networks.
\newblock In C.~Cortes, N.~D. Lawrence, D.~D. Lee, M.~Sugiyama, and R.~Garnett,
  editors, {\em Advances in Neural Information Processing Systems 28}, pages
  442--450. Curran Associates, Inc., 2015.

\bibitem{Castellana2019}
Daniele Castellana and Davide Bacciu.
\newblock {Bayesian Tensor Factorisation for Bottom-up Hidden Tree Markov
  Models}.
\newblock In {\em 2019 International Joint Conference on Neural Networks
  (IJCNN)}, pages 1--8. IEEE, 7 2019.

\bibitem{Cichocki2017TensorPerspectives}
Andrzej Cichocki, Anh~Huy Phan, Qibin Zhao, Namgil Lee, Ivan Oseledets, Masashi
  Sugiyama, and Danilo Mandic.
\newblock {Tensor networks for dimensionality reduction and large-scale
  optimizations: Part 2 applications and future perspectives}.
\newblock {\em Foundations and Trends in Machine Learning}, 9(6):431--673,
  2017.

\bibitem{Kolda2009}
Tamara~G. Kolda and Brett~W. Bader.
\newblock {Tensor Decompositions and Applications}.
\newblock {\em SIAM Review}, 51(3):455--500, 2009.

\bibitem{surv2019}
Y.~{Ji}, Q.~{Wang}, X.~{Li}, and J.~{Liu}.
\newblock A survey on tensor techniques and applications in machine learning.
\newblock {\em IEEE Access}, 7:162950--162990, 2019.

\bibitem{cpd}
Henk A.~L. Kiers.
\newblock Towards a standardized notation and terminology in multiway analysis.
\newblock {\em Journal of Chemometrics}, 14(3):105--122, 2000.

\bibitem{Tucker1966}
Ledyard~R Tucker.
\newblock {Some mathematical notes on three-mode factor analysis}.
\newblock {\em Psychometrika}, 31(3):279--311, 1966.

\bibitem{Oseledets2011}
Ivan~V. Oseledets.
\newblock {Tensor-Train Decomposition}.
\newblock {\em SIAM Journal on Scientific Computing}, 33(5):2295--2317, 1 2011.

\bibitem{tempCP}
Daniel~M. Dunlavy, Tamara~G. Kolda, and Evrim Acar.
\newblock Temporal link prediction using matrix and tensor factorizations.
\newblock {\em ACM Trans. Knowl. Discov. Data}, 5(2), February 2011.

\bibitem{anomalCP}
Evangelos Papalexakis, Konstantinos Pelechrinis, and Christos Faloutsos.
\newblock Spotting misbehaviors in location-based social networks using
  tensors.
\newblock In {\em Proceedings of the 23rd International Conference on World
  Wide Web}, pages 551--552, New York, NY, USA, 2014. Association for Computing
  Machinery.

\bibitem{session1}
Swarup~Ranjan Behera and Vijaya Saradhi.
\newblock Mining temporal changes in strengths and weaknesses of cricket
  players using tensor decomposition.
\newblock In {\em Proceedings of the European Symposium on Artificial Neural
  Networks, Computational Intelligence and Machine Learning (ESANN'20)}, 2020.

\bibitem{Weber2017EventRW}
Noah Weber, Niranjan Balasubramanian, and Nathanael Chambers.
\newblock Event representations with tensor-based compositions.
\newblock In {\em AAAI}, 2017.

\bibitem{relat1}
Maximilian Nickel, Volker Tresp, and Hans-Peter Kriegel.
\newblock A three-way model for collective learning on multi-relational data.
\newblock In {\em Proceedings of the 28th International Conference on
  International Conference on Machine Learning}, ICML'11, pages 809--816,
  Madison, WI, USA, 2011. Omnipress.

\bibitem{knowlgraph1}
Maximilian Nickel, Volker Tresp, and Hans-Peter Kriegel.
\newblock Factorizing yago: Scalable machine learning for linked data.
\newblock In {\em Proceedings of the 21st International Conference on World
  Wide Web}, WWW'12, pages 271--280, New York, NY, USA, 2012. Association for
  Computing Machinery.

\bibitem{knowlgraph2}
A.~{Padia}, K.~{Kalpakis}, and T.~{Finin}.
\newblock Inferring relations in knowledge graphs with tensor decompositions.
\newblock In {\em 2016 IEEE International Conference on Big Data (Big Data)},
  pages 4020--4022, Dec 2016.

\bibitem{ben2017mutan}
Hedi Ben-Younes, R{\'e}mi Cadene, Matthieu Cord, and Nicolas Thome.
\newblock Mutan: Multimodal tucker fusion for visual question answering.
\newblock In {\em Proceedings of the IEEE international conference on computer
  vision}, pages 2612--2620, 2017.

\bibitem{rbmTN}
Jing Chen, Song Cheng, Haidong Xie, Lei Wang, and Tao Xiang.
\newblock Equivalence of restricted boltzmann machines and tensor network
  states.
\newblock {\em Phys. Rev. B}, 97, Feb 2018.

\bibitem{pmlr-v48-cohenb16}
Nadav Cohen and Amnon Shashua.
\newblock Convolutional rectifier networks as generalized tensor
  decompositions.
\newblock In Maria~Florina Balcan and Kilian~Q. Weinberger, editors, {\em
  Proceedings of The 33rd International Conference on Machine Learning},
  volume~48 of {\em Proceedings of Machine Learning Research}, pages 955--963,
  New York, New York, USA, 20--22 Jun 2016. PMLR.

\bibitem{indbiasTN}
Nadav Cohen and Amnon Shashua.
\newblock Inductive bias of deep convolutional networks through pooling
  geometry.
\newblock In {\em Proceedings of The International Conference on Learning
  Representations (ICLR 2017)}, 2017.

\bibitem{DBLP:Calvi}
Giuseppe~Giovanni Calvi, Ahmad Moniri, Mahmoud Mahfouz, Zeyang Yu, Qibin Zhao,
  and Danilo~P. Mandic.
\newblock Tucker tensor layer in fully connected neural networks.
\newblock {\em CoRR}, abs/1903.06133, 2019.

\bibitem{yang2017deep}
Yongxin Yang and Timothy~M. Hospedales.
\newblock Deep multi-task representation learning: A tensor factorisation
  approach.
\newblock In {\em International Conference on Learning Representations (ICLR)},
  2017.

\bibitem{Frasconi1998}
Paolo Frasconi, Marco Gori, and Alessandro Sperduti.
\newblock {A general framework for adaptive processing of data structures}.
\newblock {\em IEEE Transactions on Neural Networks}, 9(5):768--786, 1998.

\bibitem{Bacciu2012}
Davide Bacciu, Alessio Micheli, and Alessandro Sperduti.
\newblock {Compositional Generative Mapping for Tree-Structured Data - Part I:
  Bottom-Up Probabilistic Modeling of Trees}.
\newblock {\em IEEE Transactions on Neural Networks and Learning Systems},
  23(12):1987--2002, 2012.

\bibitem{Bacciu2013}
Davide Bacciu, Alessio Micheli, and Alessandro Sperduti.
\newblock {An input-output hidden Markov model for tree transductions}.
\newblock {\em Neurocomputing}, 112:34--46, 2013.

\bibitem{Bacciu2013CompositionalModel}
Davide Bacciu, Alessio Micheli, and Alessandro Sperduti.
\newblock {Compositional Generative Mapping for Tree-Structured Data - Part II:
  Topographic Projection Model}.
\newblock {\em IEEE Transactions on Neural Networks and Learning Systems},
  2013.

\bibitem{Socher2013}
Richard Socher, Alex Perelygin, and Jy~Wu.
\newblock {Recursive deep models for semantic compositionality over a sentiment
  treebank (RNTN)}.
\newblock {\em Proceedings of the 2013 Conference on Empirical Methods in
  Natural Language Processing}, pages 1631--1642, 2013.

\bibitem{session2}
Daniele Castellana and Davide Bacciu.
\newblock Tensor decompositions in recursive neural networks for
  tree-structured data.
\newblock In {\em Proceedings of the European Symposium on Artificial Neural
  Networks, Computational Intelligence and Machine Learning (ESANN'20)}, 2020.

\bibitem{tuckerRecursive2020}
Daniele Castellana and Davide Bacciu.
\newblock Generalising recursive neural models by tensor decomposition.
\newblock Submitted, 2020.

\bibitem{soattoSGD}
P.~{Chaudhari} and S.~{Soatto}.
\newblock Stochastic gradient descent performs variational inference, converges
  to limit cycles for deep networks.
\newblock In {\em 2018 Information Theory and Applications Workshop (ITA)},
  pages 1--10, Feb 2018.

\bibitem{bacciu2019gentle}
Davide Bacciu, Federico Errica, Alessio Micheli, and Marco Podda.
\newblock A gentle introduction to deep learning for graphs, 2019.

\bibitem{cgmm}
Davide Bacciu, Federico Errica, and Alessio Micheli.
\newblock Contextual graph {M}arkov model: A deep and generative approach to
  graph processing.
\newblock In {\em Proceedings of the 35th International Conference on Machine
  Learning}, volume~80, pages 294--303. PMLR, 2018.

\bibitem{iclr19}
Federico Errica, Marco Podda, Davide Bacciu, and Alessio Micheli.
\newblock A fair comparison of graph neural networks for graph classification.
\newblock In {\em Proceedings of the Eighth International Conference on
  Learning Representations (ICLR 2020)}, Apr 2020.

\bibitem{how-powerful-gnn}
Keyulu Xu, Weihua Hu, Jure Leskovec, and Stefanie Jegelka.
\newblock How powerful are graph neural networks?
\newblock {\em International Conference on Learning Representations (ICLR)},
  2019.

\bibitem{esann20Errica}
Federico Errica, Davide Bacciu, and Alessio Micheli.
\newblock Theoretically expressive and edge-aware graph learning.
\newblock In Michel Verleysen, editor, {\em Proceedings of the European
  Symposium on Artificial Neural Networks, Computational Intelligence and
  Machine Learning (ESANN'20)}, Apr 2020.

\bibitem{scikit}
Maximilian Nickel and Evert Rol.
\newblock {\tt scikit-tensor} - a {Python} module for multilinear algebra and
  tensor factorizations.
\newblock pypi.org/project/scikit-tensor-py3/, 2016.

\bibitem{tensord}
Liyang Hao, Siqi Liang, Jinmian Ye, and Zenglin Xu.
\newblock Tensord: A tensor decomposition library in tensorflow.
\newblock {\em Neurocomputing}, 318:196 -- 200, 2018.

\bibitem{tensorly}
Jean Kossaifi, Yannis Panagakis, Anima Anandkumar, and Maja Pantic.
\newblock Tensorly: Tensor learning in python.
\newblock {\em Journal of Machine Learning Research}, 20(26):1--6, 2019.

\bibitem{hottbox}
Ilya Kisil, Giuseppe~G. Calvi, Bruno~Scalzo Dees, and Danilo~P. Mandic.
\newblock {HOTTBOX: Higher Order Tensors ToolBOX}.
\newblock {\em Under review}, 2020.

\end{thebibliography}
\end{document}